\documentclass[letterpaper,10pt,conference]{ieeeconf}

\IEEEoverridecommandlockouts                              

\usepackage{graphics} 
\usepackage{epsfig} 
\usepackage{listings}
\usepackage{color}
\usepackage{xcolor,colortbl}
\usepackage{tabularx}
\usepackage{cellspace}
\usepackage[Export]{adjustbox}
\usepackage{subfig}
\usepackage{lipsum}
\usepackage{amsfonts}
\usepackage{subcaption}
\usepackage{tikz}
\usepackage{pgfplots}
\usepackage{graphicx}

\usepackage{caption}
\newsavebox{\arrangebox}

\definecolor{mypink1}{rgb}{0.78, 0.21, 0.13}

\usepackage[
    pagebackref,
    breaklinks,
    colorlinks  
]{hyperref}

\hypersetup{
    colorlinks=True,
    linkcolor=mypink1,
    filecolor=magenta
}
\usepackage{cleveref}

\usepackage{booktabs}
\usepackage{balance}
\usepackage{multirow}
\usepackage{overpic}
\usepackage{cite}
\usepackage[font=small,labelfont=bf]{caption}
\usepackage{color}
\usepackage{pifont}
\makeatletter
\@namedef{ver@everyshi.sty}{}
\makeatother
\usepackage{tikz}
\usetikzlibrary{patterns}

\usepackage{colortbl}
\usepackage{pgfplots}
\pgfplotsset{compat=1.17}
\usepackage{tabularx}
\usepackage{arydshln}
\usepackage{amssymb}%
\usepackage{pifont}%

\usepackage{wasysym}%
\let\labelindent\relax
\usepackage{enumitem}
\usepackage{wrapfig}
\usepackage{xspace}
\usepackage{tabu}
\usepackage[super]{nth}
\usepackage{scalefnt}
\usepackage{dsfont}
\usepackage{graphicx,calc}
\usepackage{subcaption}
\usepackage{algorithm}
\usepackage{algpseudocode}

\definecolor{darkgreen}{RGB}{0,153,51}
\definecolor{linkgreen}{RGB}{52,130,48}
\definecolor{LightCyan}{rgb}{0.87,0.92,0.96}
\definecolor{m_green}{RGB}{233, 254, 187}
\definecolor{m_orange}{RGB}{255, 212, 121}
\definecolor{m_red}{RGB}{255, 190, 188}
\definecolor{m_violet}{RGB}{215, 131, 255}
\definecolor{m_blue}{RGB}{186, 234, 255}
\definecolor{m_brown}{RGB}{255,212,120}
\definecolor{m_lightgreen}{RGB}{212,251,122}
\definecolor{notetext}{rgb}{0.7,0,0}

\newcommand{\project}{SGAligner++}

\definecolor{first}{RGB}{119, 160, 249}
\definecolor{second}{RGB}{209, 224, 255}

\definecolor{new_red}{RGB}{255, 144, 144}
\definecolor{new_green}{RGB}{20, 181, 50}
\title{\LARGE \bf
\project{}: Cross-Modal Language-Aided 3D Scene Graph Alignment
}

\begin{document}
\author{
  \vspace{-5pt}
  \begin{minipage}[t]{0.33\textwidth}
    \centering
    \textbf{Binod Singh}{*} \\
    \normalfont Technical University of Munich
  \end{minipage}
  \hfill
  \begin{minipage}[t]{0.33\textwidth}
    \centering
    \textbf{  Sayan Deb Sarkar}{*} \\
    \normalfont Stanford University
  \end{minipage}
  \hfill
  \begin{minipage}[t]{0.33\textwidth}
    \centering
    \textbf{Iro Armeni} \\
    Stanford University
  \end{minipage} \\ \\
  {\href{https://singhbino3d.github.io/sgpp/}{sgalignerpp.github.io}}
  \thanks{\noindent\rule{0.4\textwidth}{0.4pt}
\newline \hspace*{2em}$^*$Equal Contribution.}
}

\maketitle
\thispagestyle{empty}
\pagestyle{empty}

\begin{abstract}
Aligning 3D scene graphs is a crucial initial step for several applications in robot navigation and embodied perception. Current methods in 3D scene graph alignment often rely on single-modality point cloud data and struggle with incomplete or noisy input. We introduce \textit{\project{}}, a cross-modal, language-aided framework for 3D scene graph alignment. Our method addresses the challenge of aligning partially overlapping scene observations across heterogeneous modalities by learning a unified joint embedding space, enabling accurate alignment even under low-overlap conditions and sensor noise. By employing lightweight unimodal encoders and attention-based fusion, \project{} enhances scene understanding for tasks such as visual localization, 3D reconstruction, and navigation, while ensuring scalability and minimal computational overhead. 
Extensive evaluations on real-world datasets demonstrate that \project{} outperforms state-of-the-art methods by up to 40\% on noisy real-world reconstructions, while enabling cross-modal generalization.

\end{abstract}

\thispagestyle{empty}
\pagestyle{empty}


\section{Introduction}
3D scene understanding is a foundational challenge in robotics and computer vision~\cite{Koch_2024_WACV,hughes2022hydra} and serves as the basis for applications in mixed reality, robot navigation, and embodied perception. Recently, cross-modal approaches~\cite{sarkar2025crossover,miao2024scenegraphloc} have gained significant attention due to their ability to bridge different types of data. In particular, tasks such as robot navigation~\cite{Rosinol21}, object-centric planning~\cite{agia2022taskography}, and semantic SLAM~\cite{hughes2022hydra} require accurate and consistent scene understanding across modalities. As robots operate in dynamic, real-world environments, multimodal fusion becomes crucial. Different sensory data--visual, depth, and textual information--capture distinct aspects of the scene. Fusing these modalities into a coherent spatial-semantic map enables better perception, decision-making, and interaction.

The 3D semantic scene graph is a structured representation that captures the context of the scene as an attributed and directed graph, enabling unified spatial understanding~\cite{armeni20193d, learn3dssg, 3dsg}. Recent work has explored the alignment of 3D scene graphs to support downstream tasks such as 3D registration~\cite{sarkar2023sgaligner,sgpgm,Liu2025SGRegGA} and visual localization~\cite{miao2024scenegraphloc}. These capabilities are especially important in robot navigation, for example, when a robot scanning an environment attempts to match current observations with a prior map generated from a different sensing modality. In such cases, alignment remains highly challenging, for instance, when fusing LiDAR scans with CAD-based graphs or grounding objects in a 3D point cloud using natural language descriptions. However, existing methods~\cite{sgpgm,sarkar2023sgaligner} rely on single-modal (unimodal) sensor data, mainly point clouds, and a fixed label vocabulary for graph annotations. Thus, they struggle with incomplete reconstructions and lack the flexibility to handle text-grounded or multimodal scenes, motivating a lightweight, generalizable framework for cross-modal alignment under noise.

\begin{figure}
    \centering
    \label{fig:teaser}
    \includegraphics[width=1.0\columnwidth]{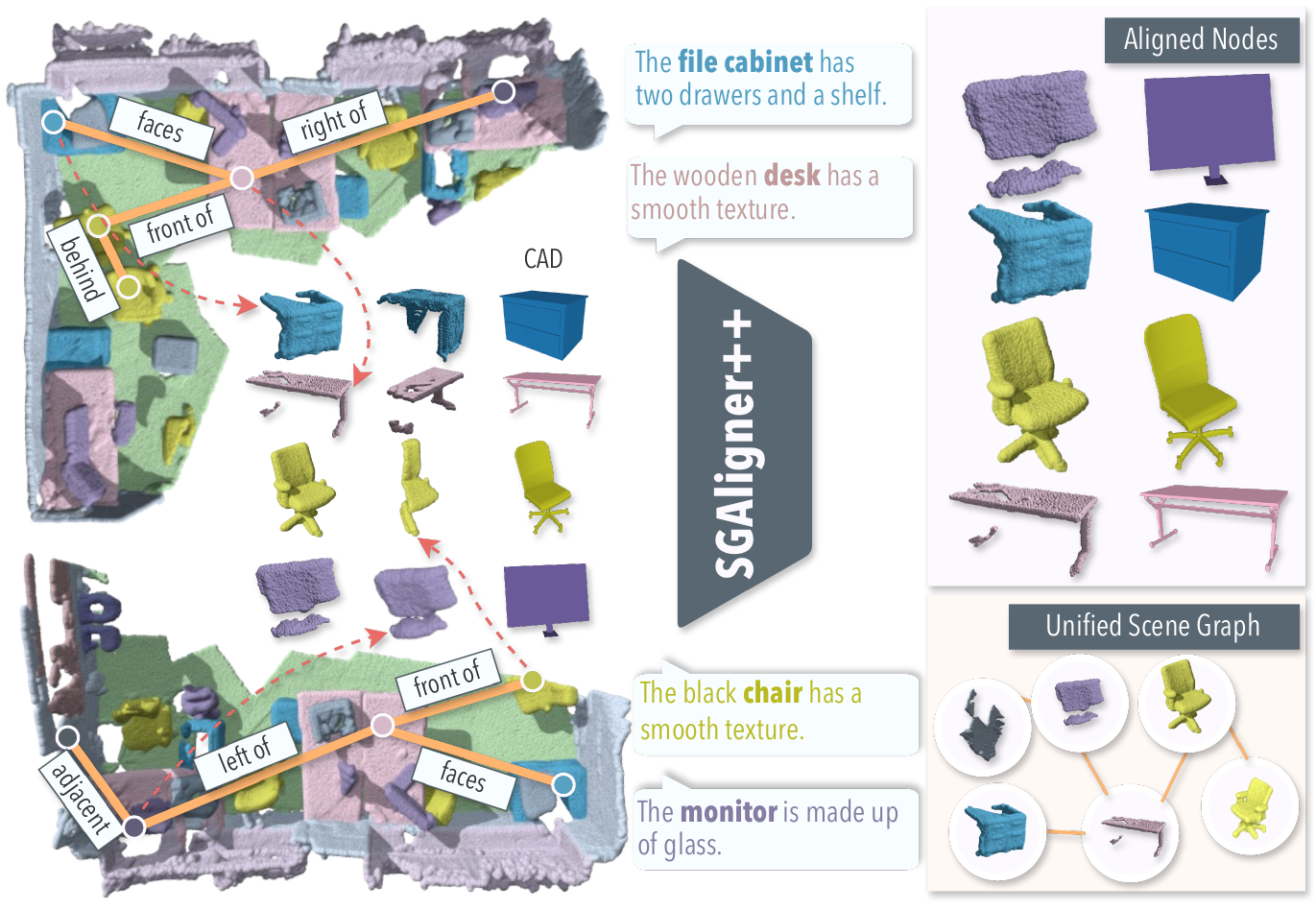}
    \caption{\textbf{\project{}.} We address the problem of aligning 3D scene graphs across different modalities, namely, point clouds, CAD meshes, text captions, and spatial referrals, using a joint embedding space. Our approach creates a unified 3D scene graph, ensuring that spatial relationships are accurately preserved. It enables robust 3D scene understanding for visual localization and robot navigation.} 
    \vspace{-22pt}
\end{figure} 

To address these challenges, we introduce \textit{\project}, a method that fuses structural, geometric, and linguistic information into a unified representation space. This enables the model to reason about spatial relationships from language, capture rich semantic context, and resolve ambiguities using multimodal data. Our approach enables efficient, scalable alignment even in noisy or low-overlap settings. Its modular design supports new modalities, and language grounding ensures adaptability to diverse 3D environments. \project{} learns a unified embedding space by representing objects through features from multiple modalities: point clouds, CAD meshes, text captions, and spatial referrals. Furthermore, \project{} handles missing data, such as when sensor inputs vary across environments (e.g., when a robot uses visual data but lacks text referrals). Our approach ensures robust object representation despite such gaps.

Unlike previous methods~\cite{sarkar2023sgaligner,sgpgm,miao2024scenegraphloc} that depend on fixed semantic labels and simplistic interpretation of object relationships in 3D scene graphs, \project{} enables seamless multimodal scene graph integration that could be used in downstream tasks such as visual localization, 3D reconstruction, and navigation.  The approach is fast, lightweight, and annotation-free, making it ideal for embodied perception. We summarize the contributions as follows:

\begin{itemize}
    \item We enable cross-modal alignment and generate a unified 3D scene graph to achieve semantic consistency.
    \item We use lightweight unimodal encoders with attention-based fusion for robust matching on noisy input.
    \item We achieve state-of-the-art performance while maintaining low runtime and memory overhead, ensuring scalable deployment.
\end{itemize}

The data and code will be made public upon acceptance.
\section{Related Work}
\noindent\textbf{Graph Matching and Partial Alignment}~\cite{MultiKE} refers to the goal of finding one-to-one alignment between graphs. EVA~\cite{eva} uses visual and auxiliary knowledge for entity alignment in both supervised and unsupervised settings. MCLEA~\cite{lin2022multi} employs multiple encoders to generate modality-specific representations, incorporating contrastive learning with intra-modal and inter-modal losses to learn cross-modal embeddings. While graph matching has been widely studied in computer vision~\cite{sarlin20superglue, miao2024scenegraphloc}, it has also proven critical in robotics applications such as mapping, localization, and SLAM~\cite{hughes2022hydra}, where establishing consistent correspondences between partial scene observations is essential. Following these advances in multi-modality knowledge graph alignment, \project{} extends the idea fusing cross-modal information to support embodied tasks such as visual localization, semantic mapping, and navigation.

\noindent\textbf{3D Scene Graph Alignment} has emerged as a core enabler for downstream tasks such as scene reconstruction, mapping, and localization.  In 3D scene graph alignment, two non-overlapping graphs are typically aligned with partial correspondences. SGAligner~\cite{sarkar2023sgaligner} first explored scene graph alignment in indoor 3D scenes and its applications such as
point cloud registration with non-overlap early stopping, point cloud mosaicking, and 3D scene alignment with changes. SG-PGM~\cite{sgpgm} shows that semantic and geometric fusion can yield high-precision alignments, though at a high computational cost. Recently, SG-Reg~\cite{Liu2025SGRegGA} introduced a GNN-based coarse-to-fine semantic scene graph registration approach. However, these methods focus on establishing pairwise correspondences between objects across graphs, without producing an integrated representation. In contrast, \project{} not only aligns nodes but also constructs a unified scene graph that merges overlapping objects and preserves intra-scene structure. This unified representation provides semantic consistency across modalities and directly supports downstream robotic tasks.

\noindent\textbf{Cross-Modal Localization and Fusion} leverages multimodal input for determining the position and orientation of an
agent within a pre-built map, pivotal across robotic applications. Where Am I~\cite{whereami} uses
an open-set natural language query to identify a scene represented by a 3D scene graph. SceneGraphLoc~\cite{miao2024scenegraphloc} goes a step ahead to address visual localization from a database of 3D scenes represented as scene graphs. Although multi-modal knowledge graph alignment~\cite{lin2022multi} has been explored in other works, the task in 3D from multiple sensor data has been largely unexplored. \project{} addresses this by demonstrating the efficacy of language-aided multimodal scene graph fusion.
\section{\project{}}
\begin{figure*}[ht!]
\vspace{10pt}
    \centering
    \includegraphics[trim=0 0 0 0,clip,width=0.9\linewidth]{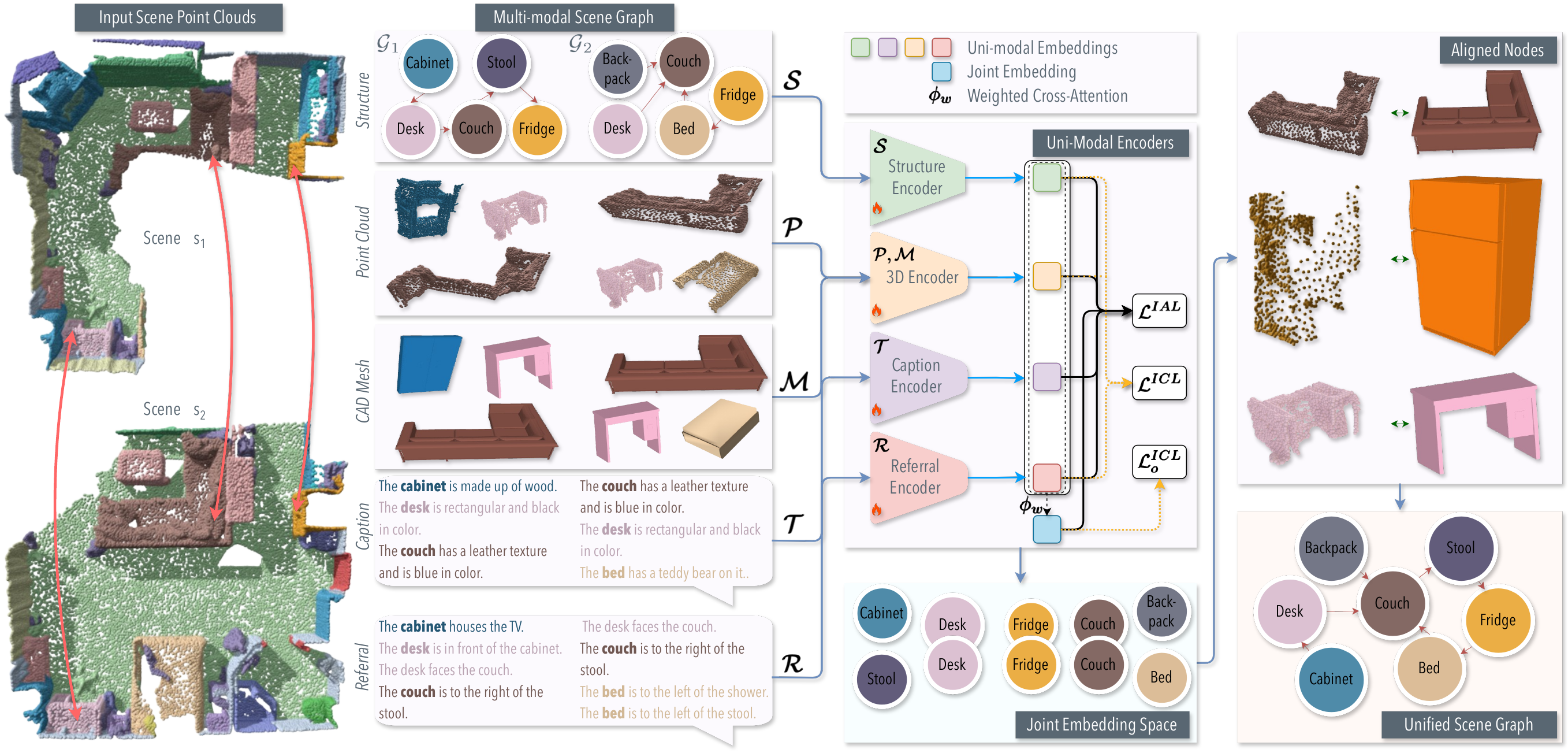}
    \begin{tabular}{cccc}
          (a) & \hspace{3cm} (b) & \hspace{4.25cm} (c) & \hspace{3cm} (d) \\
    \end{tabular}
    \caption{\textbf{Overview of \project{}.} Our method takes as input: (a) two scene point clouds with spatially overlapping objects, and (b) their corresponding 3D scene graphs with multi-modal information--point clouds, CAD meshes, text captions, and spatial referrals. (c) We process the data via separate uni-modal encoders and optimize them together in a joint embedding space using trainable attention. (d) Similar nodes are aligned together in the common space and we finally output a unified 3D scene graph, which preserves spatial-semantic consistency and enables multiple downstream tasks.} 
    \label{fig:architecture}
    \vspace{-15pt}
\end{figure*}
\subsection{Problem Formulation}
\label{sec:problem_formulation}
Let a 3D scene be $s$ and its corresponding scene graph be defined as $\mathcal{G} = (\mathcal{V}, \mathcal{E})$, where, the nodes $\{v_i\}_{i=1}^N \in \mathcal{V}$ represent \textit{object instances} and the edges $e_{jk} \in \mathcal{E}$, symbolize the \textit{relationships} between nodes $v_j$ and $v_k$ such as \textit{supported by} or \textit{standing on}.  Inspired by SceneGraphLoc~\cite{miao2024scenegraphloc}, we enrich each node $v_i$ with multiple modalities: point cloud $\mathcal{P}$, digital mesh $\mathcal{M}$, text captions $\mathcal{T}$ describing the appearance, material, and affordances~\cite{ssg3d,jia2024sceneversescaling3dvisionlanguage}, referrals $\mathcal{R}$ describing spatial context within a scene, and a structure graph $\mathcal{S}$ encoding the object layout. These modalities integrate multimodal information directly into the scene graph, enhancing its representation. Our setup can be easily extended to more modalities, such as depth maps or surface normals. Similarly to~\cite{sarkar2023sgaligner,sgpgm}, given two scene graphs, $\mathcal{G}_1 = (\mathcal{V}_1, \mathcal{E}_1)$ and $\mathcal{G}_2 = (\mathcal{V}_2, \mathcal{E}_2)$, which represent partial observations of the same scene, we aim to construct a unified scene graph $\mathcal{G}^*$ that preserves spatial-semantic consistency while integrating cross-scene correspondences. This involves handling partial overlap and sensor noise, optimizing the objective function:

\begin{equation}
    \arg\max_{\textbf{S}}
    f(\textbf{S};\mathcal{G}_1, \mathcal{G}_2),
\end{equation}

\noindent in which $\textbf{S} \in [0,1]^{N_1  \times N_2 }$ is the pairwise cosine similarity matrix that maps nodes between the source graph $\mathcal{G}_1$ and the reference graph $\mathcal{G}_2$. To achieve 3D scene graph alignment, our goal is to create a unified joint embedding space by integrating the $K$ modalities of each vertex $v_i$ through multi-modal knowledge distillation, ensuring consistency across all modalities. In the absence of annotations, we feed instance-based \textit{referrals} to a context-aware LLM for 3D scene graph generation (more details in Sec~\ref{sec:experiment}).

\subsection{3D Scene Graph Alignment}
\label{sec:scene_graph_alignment}
Similar to SGAligner~\cite{sarkar2023sgaligner}, \project{} uses separate unimodal encoders to extract features for each of the $K$ modalities and the structure graph $\mathcal{S}$. These embeddings are jointly optimized in a shared latent space to generate a cohesive and aligned scene graph representation. An overview of our method is in Fig. \ref{fig:architecture}.

\noindent \textbf{Point Cloud and Mesh.} For each instance $v_i$, we process its associated point cloud $\mathcal{P}_i$ and mesh $\mathcal{M}_i$ using a PointNet backbone~\cite{qi2016pointnet} to extract geometric features $\phi_i^\mathcal{P}$ and $\phi_i^\mathcal{M}$ respectively. Notably, we intentionally exclude semantic class information during feature extraction to achieve a semantically agnostic feature space. $\mathcal{P}_i$ is uniformly downsampled to $K=512$ points by farthest point sampling. For $\mathcal{M}_i$, we first convert the mesh to a point cloud using uniform surface area sampling and then subsample $2048$~\cite{sarkar2025crossover} points as input to the encoder. Our experiments demonstrate that \project{} maintains robust performance even with reduced $K$, highlighting its suitability for resource-constrained applications.

\noindent \textbf{Structure Graph.} To capture spatial relationships and object layout in a scan, we leverage the 3D scene graph using a \textit{structure graph} representation. We define node features as the centroid and extent of the 3D bounding box of the object $v_i$ and the \textit{referrals} make up the edges. Unlike~\cite{sarkar2023sgaligner}, our node features explicitly model object geometry. A two-layer Graph Attention Network (GAT)~\cite{veličković2018graphattentionnetworks} with a diagonal weight matrix aggregates multi-hop neighborhood information, producing structure embeddings $\phi_i^\mathcal{S}$ for efficient reasoning.

\noindent \textbf{Text Captions and Referrals.} Text captions $t_i$ contain information about objects in a scene, such as ``\textit{The curtain is made of green taffeta material and has a triangular shape}''. We encode such captions with a text encoder~\cite{blip2} to form features $\phi_i^\mathcal{T}$. To describe the scene context from the point of view of a single object $v_i$, we collect all $r_i$ referrals for each object, encode via the same encoder, and aggregate them via average grouping to form the referral feature vector $\phi_i^\mathcal{R}$.

\noindent \textbf{Multi-Modal Fusion.} Once we have created uni-modal instance-wise features, we optimize them together in a joint embedding space. Uni-modal features $\phi_i^k$ for $k \in \mathcal{K} = \{\mathcal{P}, \mathcal{M}, \mathcal{S}, \mathcal{T}, \mathcal{R}\}$ are passed via a projection head and fused using trainable attention weights $w_k$:  
\begin{equation}
    \hat\phi_i = MLP \left (\sum_{k \in \mathcal{K}} \frac{\exp(w_k)}{\sum_{j} \exp(w_j)} \phi_i^k \right)
\end{equation}  
where $L_2$-normalized features are distilled to joint embeddings $\hat\phi_i$. Following~\cite{lin2022multi,sarkar2023sgaligner} we optimize our alignment using \textit{Intra-Modal Contrastive Loss (ICL)} to separate embeddings of non-matching objects within each modality and \textit{Inter-Modal Alignment Loss (IAL)} to align matched objects across modalities. The total loss is:  
\begin{equation}
    \mathcal{L} = \mathcal{L}_o^{ICL} + \sum\nolimits_{k \in \mathcal{K}} (\alpha_k \mathcal{L}_k^{ICL} + \beta_k \mathcal{L}_k^{IAL}),
\end{equation}  
where $o$ refers to the joint embedding $\hat\phi_i$, and $k \in \mathcal{K}$ refers to the uni-modal embeddings  $\phi_i^k$. $\alpha_k$ and $\beta_k$ are the parameters learned within a multi-task learning paradigm using homoscedastic uncertainty~\cite{homeomulti}.

\noindent \textbf{Creating A Unified Scene Graph.} In the joint embedding space,
similar nodes are located closely. We use cosine similarity to first match nodes. The matched set of object nodes corresponds to spatially overlapping regions of the
two scenes, $F = \{(v_1, v_2) | v_1 \equiv
v_2, v_1 \in \mathcal{V}_1, v_2 \in \mathcal{V}_2 \}$. Using these nodes, we construct a unified scene graph $\mathcal{G}^*$ by merging their attributes and fusing multimodal data such as point clouds $\mathcal{P}$ and meshes $\mathcal{M}$. Unmatched nodes keep their original connectivity, yielding a unified representation that preserves intra-scene structure while ensuring semantic alignment across overlapping scenes.
\section{Data}
\label{sec:data}

We train and evaluate \project{} on ScanNet~\cite{scannet} and 3RScan~\cite{3rscan} datasets. For CAD mesh annotations on ScanNet, we use the ScanNotate~\cite{scannotate} dataset, which is based on a Monte Carlo Tree Search approach for retrieval of pose-aligned CAD models from RGB-D scans. We leverage \textit{captions} and \textit{referrals} from SceneVerse~\cite{jia2024sceneversescaling3dvisionlanguage}, a 3D vision-language dataset with 2.5M pairs. Following the sub-scan generation in~\cite{sarkar2023sgaligner} for ground truth data with $10-90$\% overlap, we generate: (1) $5510$ training sub-scans forming $7784$ pairs and $1427$ validation subscans forming $2024$ pairs, for ScanNet; (2) $5500$ training subscans forming $10890$ pairs and $729$ validation scans forming $1449$ pairs, for 3RScan. In the absence of explicit scene graph annotations, we construct them automatically using \textit{referrals} (Sec.~\ref{sec:scene_graph_gen}). To showcase robustness against on-the-fly 3D reconstructions and scene graphs, we evaluate our method on predicted data, the generation of which is detailed in Sec.~\ref{sec:pred_data_gen}. 

\subsection{Scene Graph Generation}
\label{sec:scene_graph_gen}

\begin{figure}[h!]
\vspace{-5pt}
    \centering
    \includegraphics[width=\linewidth]{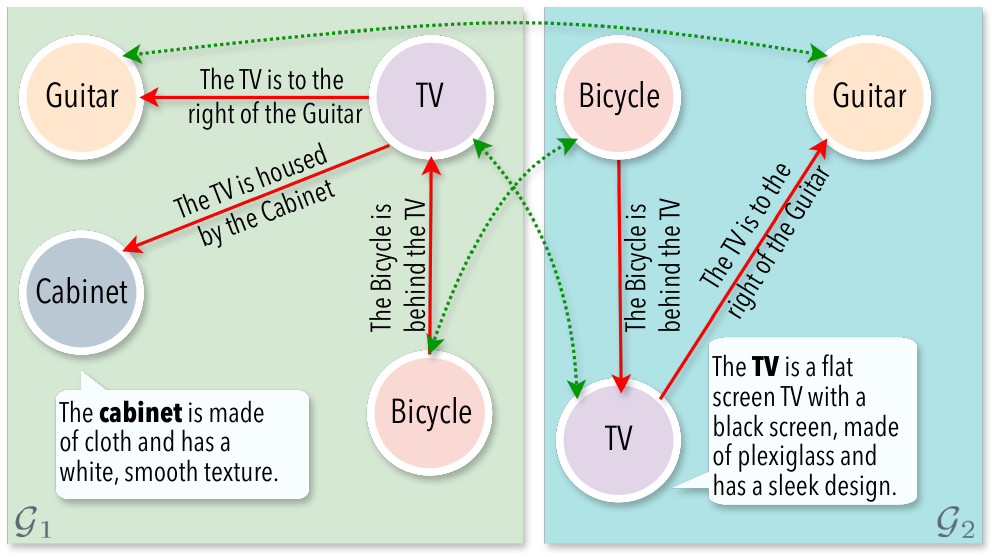}
    \caption{\textbf{Example of context-aware LLM-generated scene graphs}. Overlapping pairs are in \textcolor{green!50!black}{green} and non-overlapping are in \textcolor{red}{red}.}
    \label{fig:scene_graph_gen}
\vspace{-16pt}
\end{figure}

For 3RScan~\cite{3rscan}, we use ground truth scene graphs from 3DSSG~\cite{ssg3d}. For ScanNet~\cite{scannet}, which lacks scene graphs, we generate $\mathcal{S}$ directly using \textit{referrals} from SceneVerse~\cite{jia2024sceneversescaling3dvisionlanguage} with the following LLM-based procedure. We employ an LLM (GPT-4o-mini) to construct the scene graph $\mathcal{G} = (\mathcal{V}, \mathcal{E})$, where each object $v_i \in \mathcal{V}$ is described by a caption $t_i$ and a set of spatial referrals $\mathcal{R}$, alongside the other modalities, point cloud $\mathcal{P}$ and mesh $\mathcal{M}$. Each referral $r_j = (v_i, \tau_j)$ expresses a spatial relation between a primary object $v_i$ and an unspecified secondary object $\tau_j$. The LLM serves as a disambiguation function, jointly reasoning over $(r_i, t_i, \mathcal{R})$, to identify candidate secondary objects. An edge is added only when a unique candidate exceeds a confidence threshold, leading to a sparse and unambiguous graph. The resulting $\mathcal{G}$ captures spatial-semantic relationships by combining geometric structure with open-vocabulary semantics, enabling hierarchical and relational reasoning, as shown in Fig.~\ref{fig:scene_graph_gen}.
 
\definecolor{maroonbrown}{RGB}{102, 28, 28} 

\subsection{Predicted Data Generation} 
\label{sec:pred_data_gen}

For deployment in real-world scenarios, methods must robustly handle predicted noisy and semantically imperfect data. We utilize SceneGraphFusion~\cite{wu2021scenegraphfusionincremental3dscene} to obtain 3D reconstructions and scene graph prediction. Unlike ground truth data, which relies on precise instance segmentation, SceneGraphFusion performs geometric segmentation, leading to object instances that may not perfectly align with annotated labels. To enable evaluation, we establish correspondences between objects in the ground truth and prediction by comparing their bounding boxes. Specifically, for object point clouds $\mathcal{P}_i$, we match the nearest bounding boxes and propagate ground truth annotations to the predicted data. This approach ensures annotation consistency while accommodating segmentation discrepancies. Similarly to ground truth, using the validation set, we generate $620$ scans that form $1241$ pairs on 3RScan and $1520$ forming $2991$ pairs on ScanNet, for evaluation.
\begin{table}[h]
\centering
\setlength{\tabcolsep}{2pt}
\caption{\textbf{Node Matching on 3RScan with Ground Truth and Predicted Data.} \project{} outperforms baselines in both cases and shows the most improvement over them in the predicted one.}
\label{tab:alignment_scan3r}
\begin{tabular}{lcccc}
\toprule
& & \multicolumn{3}{c}{\textbf{Hits@} $\uparrow$} \\
\textbf{Model} & \textbf{Mean RR $\uparrow$} & \textbf{K = 1} & \textbf{K = 3} & \textbf{K = 5} \\
\midrule
\multicolumn{5}{l}{\textcolor{black!60}{\textit{Predicted 3D Reconstruction + Scene Graphs}}} \\[-0.2ex] %
\textcolor{black!100}{EVA~\cite{eva}} & 33.39 & 17.74  &  37.85 & 51.93 \\ 
\textcolor{black!100}{SGAligner~\cite{sarkar2023sgaligner}} & 34.09 & 18.57 & 38.35 & 52.09   \\ 
\textcolor{black!100}{SGAligner++} & \textbf{69.01} & \textbf{55.42} & \textbf{79.25} & \textbf{87.78} \\ 

&\textcolor{darkgreen}{\scriptsize(+\,$34.92$)} & \textcolor{darkgreen}{\scriptsize(+\,$36.85$)} & \textcolor{darkgreen}{\scriptsize(+\,$40.90$)}
& \textcolor{darkgreen}{\scriptsize(+\,$35.69$)} \\

\arrayrulecolor{black!10}\midrule\arrayrulecolor{black}

\multicolumn{5}{l}{\textcolor{black!60}{\textit{Ground Truth Mesh + Annotated Scene Graphs}}} \\[-0.2ex] %
\textcolor{black!100}{EVA~\cite{eva}} & 75.36 & 64.28 & 83.83 & 90.66 \\ 
\textcolor{black!100}{SGAligner~\cite{sarkar2023sgaligner}} & 77.81 & 66.24 & 87.86 & 93.89 \\ 
\textcolor{black!100}{SG-PGM~\cite{sgpgm}} & 96.94 & 94.63 & 99.23  & 99.82  \\ 
\textcolor{black!100}{SGAligner++} & \textbf{99.43} & \textbf{99.06} & \textbf{99.79} & \textbf{99.89}  \\ 
&\textcolor{darkgreen}{\scriptsize(+\,$2.49$)} & \textcolor{darkgreen}{\scriptsize(+\,$4.43$)} & \textcolor{darkgreen}{\scriptsize(+\,$0.56$)}
& \textcolor{darkgreen}{\scriptsize(+\,$0.07$)} \\

\bottomrule
\end{tabular}
\vspace{-10pt}
\end{table}
\section{Experiment}
\label{sec:experiment}
\par \noindent \textbf{Evaluation.} To assess performance, we evaluate \project{} on 3D scene graph alignment (Sec.~\ref{sec:scene_graph_alignment}) and analyze runtime complexity (Sec.~\ref{sec:runtime_complexity}). We compare against three state-of-the-art methods: SGAligner~\cite{sarkar2023sgaligner}, EVA~\cite{eva}, and SG-PGM~\cite{sgpgm}. For fair comparison, we re-train SGAligner and EVA using the same setting as SG-PGM~\cite{sgpgm}. Since Sceneverse~\cite{jia2024sceneversescaling3dvisionlanguage} lacks annotations for all scans, we evaluate all methods on our validation set. Additionally, we ablate cross-modal alignment (Sec.~\ref{sec:cross_modal_abl}) to validate our design choice.

\par \noindent \textbf{Implementation Details.} Following~\cite{lin2022multi,sarkar2023sgaligner}, we adopt a two-layer GAT with each hidden layer having 128-dimensional units. The projection head is a three-layer MLP with LayerNorm and ReLU activation that maps each module to a common $D=512$ embedding space to obtain the joint embedding. The model is trained for 50 epochs using the AdamW optimizer and cosine annealing scheduler with a learning rate of $0.001$, a batch size of $4$, on an NVIDIA A100 GPU with $40$GB of memory.

\subsection{3D Scene Graph Alignment}
\label{sec:scene_graph_alignment}

\par
\begin{figure*}[ht!]
    \vspace{11pt}
    \centering
    \includegraphics[trim=0 0 0 0,clip,width=0.9\linewidth]{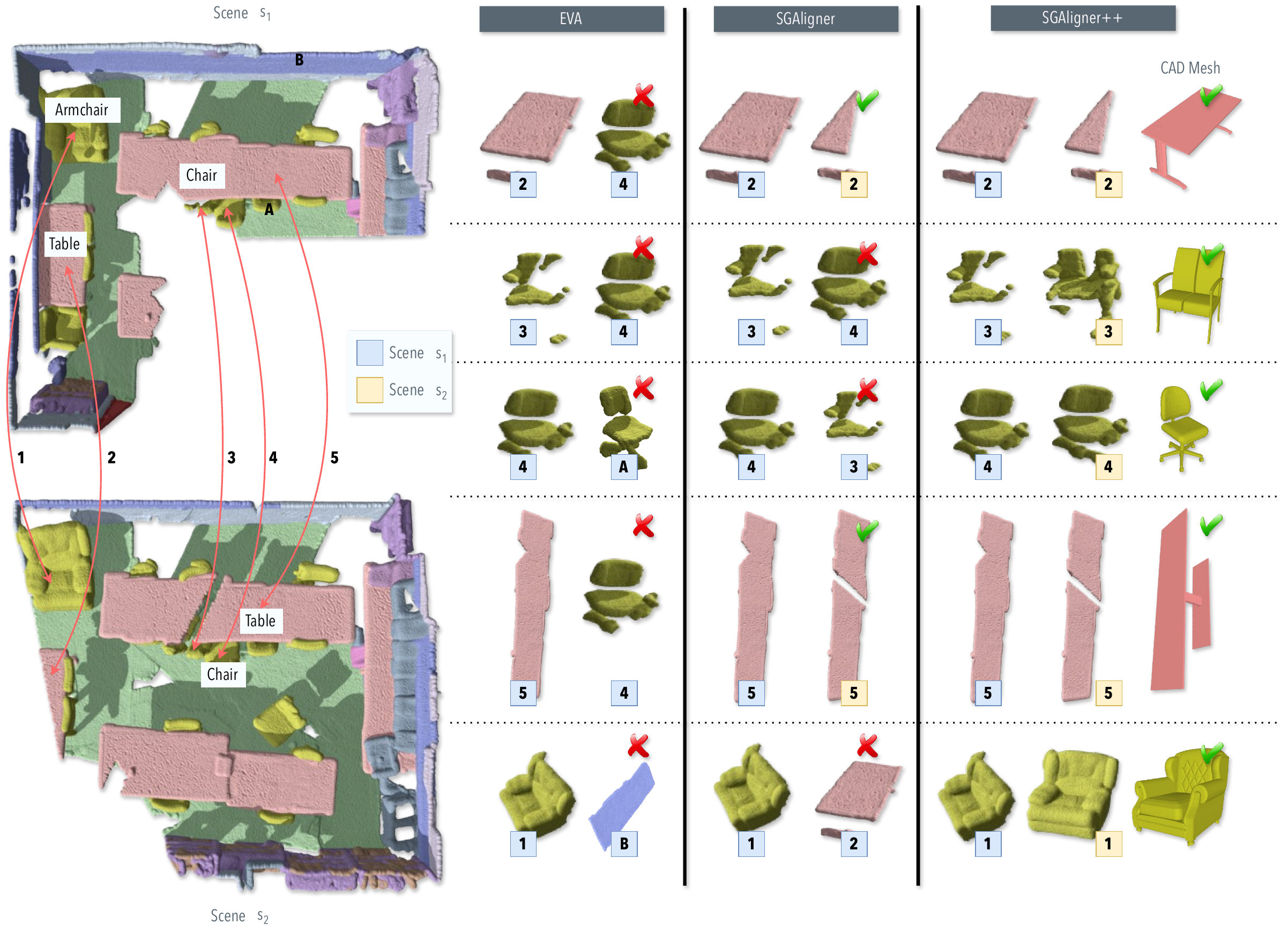}
    \caption{\textbf{Qualitative Results on Node Matching.} Given two partially overlapping observations of the same scene, EVA \cite{eva} is unable to identify any correct matches, aligning objects within the same scene and SGAligner \cite{sarkar2023sgaligner} cannot handle intra-class instances (e.g. two chairs). In contrast, \project{} correctly identifies all point cloud matches, as well as CAD ones. Numbers indicate common objects across the two overlapping scenes.}
    \label{fig:quals}
    \vspace{-9pt}
\end{figure*}

\begin{figure}[h!]
    \centering
    \includegraphics[width=\linewidth]{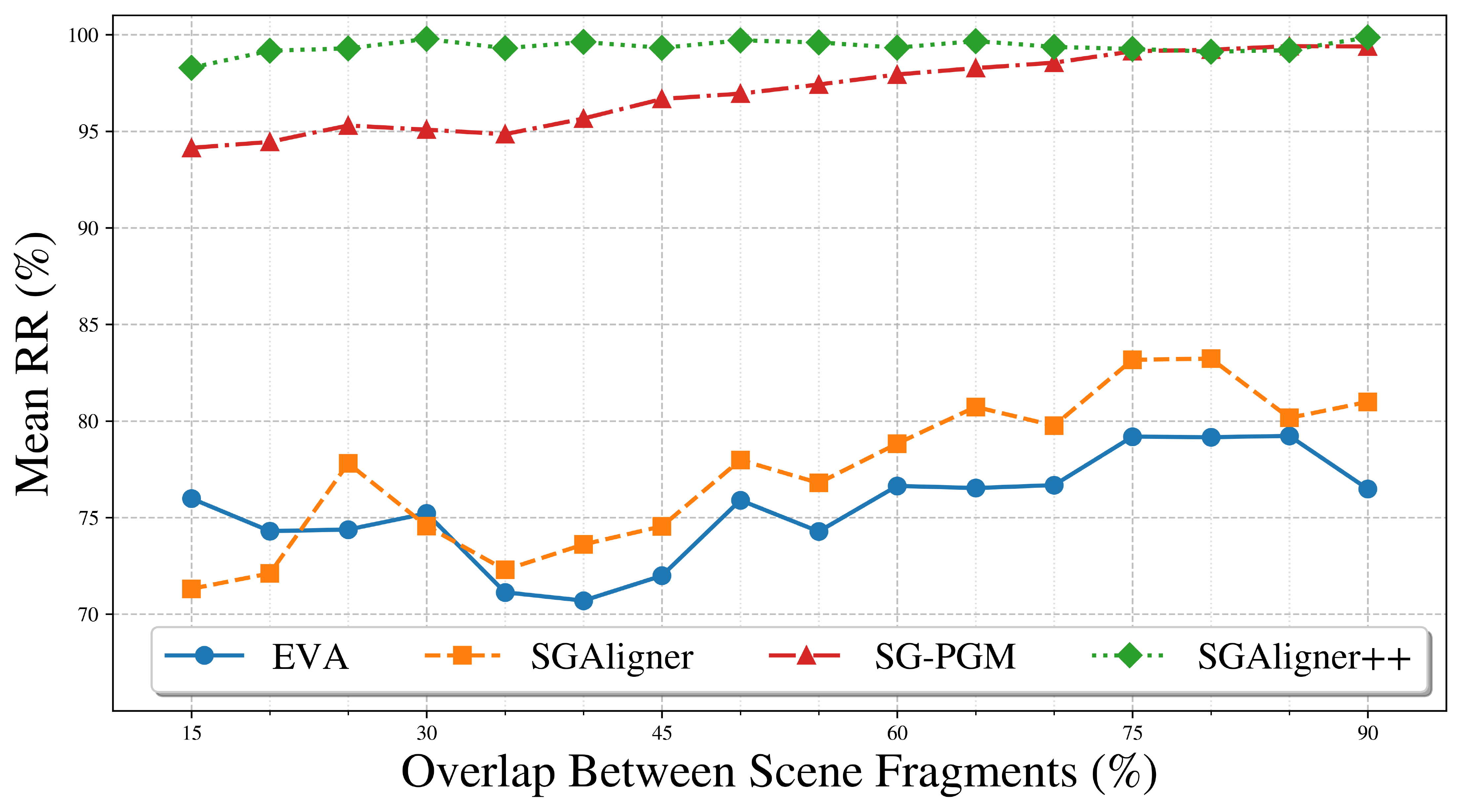}
    \caption{\textbf{Node Matching Mean RR vs. Overlap Range, on 3RScan}. \project{} generalizes across overlap thresholds and performs robustly even in low-overlap cases.}
    \label{fig:overlap_distribution_chart}
    \vspace{-12.4pt}
\end{figure}

 \noindent \textbf{Node Matching.} Following the benchmark in ~\cite{sarkar2023sgaligner} to evaluate node matching, we use Reciprocal Rank (Mean RR) and Hits@K, where K = $1,3,5$ to assess the performance of our model. MRR denotes the mean reciprocal rank of correct matches. Hits@K denotes the ratio of correct matches appearing within top K, based on their cosine similarity ranking. Aiming for a more practical evaluation~\cite{sgpgm}, we use $T \ne I_4$ by augmenting random transformation between two scene fragments. Tab.~\ref{tab:alignment_scan3r} shows results using both ground truth and \textit{predicted} data. Despite being trained on ground truth data like other methods, \project{} significantly outperforms all methods for both settings, confirming the effectiveness of our language-aided learned representation space in identifying semantically and structurally consistent correspondences. Notably, we were unable to run SG-PGM~\cite{sgpgm} on \textit{predicted data due to memory issues on a NVIDIA A100 GPU}, highlighting significant practical limitations for deployment. We show qualitative comparison in Fig.~\ref{fig:quals}. While the baseline methods often confuse furniture with supporting elements such as walls or floors, \project{} is able to give correct matches due to an efficient multimodal representation integrated with object geometry. Fig.~\ref{fig:overlap_distribution_chart} shows results divided into different scene overlapping ranges. \project{} maintains similar performance across both low- and high-overlap scene fragments, which demonstrates the importance of fusing cross-modal representations for robustness against missing data and random transformations. To further validate the sensitivity of our approach, we report performance with varying object point cloud resolution in Tab.~\ref{tab:alignment_pc_res}. Our method maintains high accuracy across point densities, demonstrating its suitability for robotic interaction tasks, particularly as an alignment module within a real-time perception system~\cite{hughes2022hydra}.

\begin{table}[h]
\centering
\setlength{\tabcolsep}{2pt}
\caption{\textbf{Node Matching on ScanNet and 3RScan with Varying Point Resolutions.} Ours is robust even in low resolution.}
\label{tab:alignment_pc_res}
\begin{tabular}{lcccc}
\toprule
& & \multicolumn{3}{c}{\textbf{Hits@} $\uparrow$} \\
\textbf{Point Resolution} & \textbf{Mean RR $\uparrow$}  & \textbf{K = 1} & \textbf{K = 3} & \textbf{K = 5} \\
\midrule
\multicolumn{5}{l}{\textcolor{black!60}{\textit{3RScan predicted data}}} \\[-0.2ex] %
\hfill $64$ & \textbf{69.16} & \textbf{55.74} & 79.17 & 87.77 \\
\hfill $128$ & 68.99  & 55.43 & 79.15 & 87.70 \\
\hfill $256$ & 69.07 & 55.54 & 79.21 & 87.64 \\
\hfill $512$ (\textit{Ours}) & 69.01 & 55.42  & \textbf{79.25} & \textbf{87.78}   \\
\arrayrulecolor{black!10}\midrule\arrayrulecolor{black}
\multicolumn{5}{l}{\textcolor{black!60}{\textit{ScanNet predicted data}}} \\[-0.2ex] %
\hfill $64$ & 75.48 & 62.70 & 86.58 & 94.02 \\
\hfill $128$ & 75.48  & 62.66 & 86.68 & 94.04    \\
\hfill $256$ & 75.48 & 62.70  & 86.57  & 93.98   \\
\hfill $512$ (\textit{Ours}) & \textbf{75.67} & \textbf{62.95}  & \textbf{86.67} & \textbf{94.14}   \\
\bottomrule
\end{tabular}
\vspace{-5pt}
\end{table}

\par \noindent \textbf{Overlap Check For Point Cloud Registration.} In practical robotic applications, determining whether two scenes overlap is often unknown a priori. Standard point cloud registration methods~\cite{geotr,predator} compute matchability scores but are typically trained and tested only on overlapping scenes, lacking mechanisms to discard non-overlapping ones. Scene-graph-based alignment methods~\cite{sarkar2023sgaligner,sgpgm} address this limitation with an alignment score, derived from the matched nodes. Our method naturally inherits this capability: by aligning scene graphs, we directly obtain a robust measure of scene overlap, without requiring additional heuristics. This enables a two-stage pipeline where we first reliably detect overlap, and then any registration method--even a brute-force one--can be applied to finalize the alignment on the filtered pairs. In Tab.~\ref{tab:overlap_check}, we report precision, recall, and F1-score for identifying overlapping scene pairs, using a validation set of 1449 overlapping and 1449 non-overlapping scans from the ground truth 3RScan dataset, under $T \ne I_4$ setting. We use scene-level \textit{alignment} score $\mathcal{\xi} \ge 0.5$ for SGAligner, EVA, and \project{}. For SG-PGM, we use \textit{matchability} score $\mu \ge 0.375$ as in their original implementation. While SG-PGM achieves a slightly higher recall, our method demonstrates superior precision and F1-score, leading to fewer incorrect matches. This significantly reduces the risk of robot navigation and interaction errors. Our results show that \project{} provides a more balanced and robust scene alignment approach, making it well-suited for real-world robotics applications where speed, accuracy, and reliability in scene registration are critical.

\begin{table}[h]
\centering
\setlength{\tabcolsep}{2pt}
\caption{\textbf{Overlap Check For Point Cloud Registration on 3RScan.} \project{} results to fewer incorrect matches than the baselines, as shown with the higher precision and F1-score.}
\label{tab:overlap_check}
\begin{tabular}{lccc}
\toprule
\textbf{Method} & \textbf{Precision $\uparrow$}  & \textbf{Recall $\uparrow$} & \textbf{F1-Score $\uparrow$} \\
\midrule

\multicolumn{4}{l}{\textcolor{black!60}{\textit{Point cloud based}}} \\[-0.2ex] %
GeoTransformer~\cite{geotr} & 99.59 & 39.90 & 56.97 \\
\arrayrulecolor{black!10}\midrule\arrayrulecolor{black}
\multicolumn{4}{l}{\textcolor{black!60}{\textit{3D Scene Graph based}}} \\[-0.2ex] %
EVA~\cite{eva} & 84.13 & 60.39 & 70.31 \\
SGAligner~\cite{sarkar2023sgaligner} & 85.37 & 79.06 & 82.09 \\
SG-PGM~\cite{sgpgm} & 65.33 & \textbf{99.86} & 78.99 \\
\project{} & \textbf{86.14} & 94.34 & \textbf{90.05} \\
&\textcolor{darkgreen}{\scriptsize(+\,$0.77$)} & \textcolor{red}{\scriptsize(-\,$5.52$)} & \textcolor{darkgreen}{\scriptsize(+\,$7.96$)} \\
\bottomrule
\end{tabular}
\vspace{-10pt}
\end{table}

\subsection{Runtime Complexity}
\label{sec:runtime_complexity}

Tab.~\ref{tab:runtime_complexity} shows the runtime and memory complexity of SG-PGM~\cite{sgpgm}, as well as each component in our pipeline. To ensure fairness and consistency, we evaluate all methods on the 3RScan ground truth validation set with values averaged across $5$ runs. Since SG-PGM directly outputs the matches, unlike our method in which the object embeddings are used to find the matched nodes, memory is not applicable in their case. Even when running inference on multiple modalities, our method does not significantly vary in runtime and memory requirements.
\begin{table}[h]
\centering
\setlength{\tabcolsep}{2pt}
\caption{{\textbf{Runtime and Computational Complexity.} Ours is significantly faster than SG-PGM and has very low memory needs.}}
\label{tab:runtime_complexity}
\begin{tabular}{lcc}
\toprule

\textbf{Method} & \textbf{Runtime (in ms)} & \textbf{Memory (in MB)} \\
\midrule

\multicolumn{3}{l}{\textcolor{black!60}{\textit{Graph + Registration Based}}} \\[-0.2ex] %
SG-PGM \cite{sgpgm} & 161.15 & 0.18 \\
\arrayrulecolor{black!10}\midrule\arrayrulecolor{black}
\multicolumn{3}{l}{\textcolor{black!60}{\textit{Ours}}} \\[-0.2ex] %
$\mathcal{P}$ & 3.92& 0.15 \\
$\mathcal{P} + \mathcal{M}$ & 5.75 & 0.22 \\
$\mathcal{P} + \mathcal{M} + \mathcal{S}$ & 10.05 & 0.29 \\
$\mathcal{P} + \mathcal{M} + \mathcal{S} + \mathcal{T}$ &24.26 & 0.36\\
$\mathcal{P} + \mathcal{M} + \mathcal{S} + \mathcal{T} + \mathcal{R}$ & 46.56  & 0.44 \\
\arrayrulecolor{black!10}\midrule\arrayrulecolor{black}
Average (\textit{Ours} - across all modules) & \textbf{18.11 $\pm$ 17.79} & \textbf{0.29 $\pm$ 0.11} \\
\bottomrule
\end{tabular}
\vspace{-7pt}
\end{table}

\subsection{Ablation Study On Cross-Modal Alignment}
\label{sec:cross_modal_abl}
To understand how language-aided 3D scene graphs can aid cross-modal understanding, we evaluate cross-modal graph alignment at the scan level (e.g, between graphs constructed from point cloud and CAD meshes), using the same metrics as defined in Sec.~\ref{sec:scene_graph_alignment}. In many robotics scenarios, sensor inputs may be incomplete or vary across environments, such as when a robot perceives a scene through depth cameras while relying on pre-existing CAD models for reference. To simulate this, we systematically vary modalities between source and reference scenes, assessing how missing information affects alignment. The results are in Tab.~\ref{tab:alignment_cross_modal_scan3r}. Please note that here we evaluate using the $T=I_4$ setting, since the focus is on understanding how the modalities correspond to each other. Our results show that \project{} remains effective in different combinations of modality, with the best performance when structural cues are preserved.

We ablate the contribution of each modality in Tab.~\ref{tab:ablation_study}, from the most common to the least. In robotic perception, each modality contributes uniquely, and adding more consistently improves alignment. On ScanNet, CAD models and referrals give the largest boost, while in 3RScan structure graphs slightly hurt performance but text captions provide a strong gain. This shows that certain modality pairs serve as stronger priors, highlighting the value of multimodal cues for robust alignment under incomplete or varying inputs.

\begin{table}[h]
\centering
\setlength{\tabcolsep}{2pt}
\caption{\textbf{Node Matching Evaluation of \project{}'s different modalities on ScanNet and 3RScan.} Each modality offers additional cues for reasoning, improving the performance. }
\label{tab:ablation_study}
\resizebox{0.45\textwidth}{!}{
\begin{tabular}{lcccc}
\toprule
& & \multicolumn{3}{c}{\textbf{Hits@} $\uparrow$} \\
\textbf{Method Component} & \textbf{Mean RR $\uparrow$} &  \textbf{K = 1} & \textbf{K = 3} & \textbf{K = 5} \\
\midrule
\multicolumn{5}{l}{\textcolor{black!60}{\textit{ScanNet predicted data}}} \\[-0.2ex] %
$\mathcal{P}$ & 75.50 & 69.06 & 78.76 & 83.43 \\
$\mathcal{P} + \mathcal{M}$ & 83.07 & 76.83 & 87.63 & 91.64 \\
$\mathcal{P} + \mathcal{M} + \mathcal{S}$ & 83.19 & 76.85 & 87.79 & 91.95 \\
$\mathcal{P} + \mathcal{M} + \mathcal{S} + \mathcal{T}$ & 86.10 & 79.72 & 91.49 & 95.30 \\
$\mathcal{P} + \mathcal{M} + \mathcal{S} + \mathcal{T} + \mathcal{R}$ & \textbf{90.45} & \textbf{84.97} & \textbf{95.58} & \textbf{98.18} \\
\arrayrulecolor{black!10}\midrule\arrayrulecolor{black}
\multicolumn{5}{l}{\textcolor{black!60}{\textit{3RScan predicted data}}} \\[-0.2ex] %
$\mathcal{P}$ & 73.35 & 67.40 & 75.80 & 80.26\\
$\mathcal{P} + \mathcal{S}$ & 66.20 & 58.81 & 69.07 & 74.48\\
$\mathcal{P} + \mathcal{S} + \mathcal{T}$ & 81.22 & 73.67 & 93.28 & 96.50 \\
$\mathcal{P} + \mathcal{S} + \mathcal{T} + \mathcal{R}$ & \textbf{87.65}  & \textbf{81.24} & \textbf{93.28} & \textbf{96.50}  \\
\bottomrule
\end{tabular}
}

\end{table}
\begin{table}[h]
\centering
\setlength{\tabcolsep}{2pt}
\vspace{4pt}
\caption{\textbf{Cross-modal evaluation of SGAligner++ at the \emph{scan level}}. Results show that \project{} aligns well across modality pairs, with best performance when structural cues are preserved, underscoring the role of geometry for cross-modal correspondence.}

\label{tab:alignment_cross_modal_scan3r}
\resizebox{0.5\textwidth}{!}{
\begin{tabular}{lcccc}
\toprule
& & \multicolumn{3}{c}{\textbf{Hits@} $\uparrow$} \\
\textbf{Module} & \textbf{Mean RR $\uparrow$}  &  \textbf{K = 1} & \textbf{K = 3} & \textbf{K = 5} \\
\midrule
\multicolumn{5}{l}{\textcolor{black!60}{\textit{ScanNet predicted data}}} \\[-0.2ex] %

$\mathcal{P} + \mathcal{M} + \mathcal{T} +  \mathcal{R}  \rightarrow   \mathcal{T} + \mathcal{R}$ & 40.43 & 23.24 & 48.11 & 62.78  \\

$\mathcal{T} +  \mathcal{R}  \rightarrow \mathcal{P} + \mathcal{M} +    \mathcal{T} + \mathcal{R}$ & 41.25 & 24.11 & 48.05 & 63.30  \\

$\mathcal{P} + \mathcal{S} + \mathcal{T} +  \mathcal{R}  \rightarrow   \mathcal{T} + \mathcal{R}$ & 42.32 & 24.92 & 51.13 & 64.58  \\

$\mathcal{P} + \mathcal{T} + \mathcal{R}  \rightarrow   \mathcal{T} + \mathcal{R}$ & 57.50 & 39.24 & 69.79 & 83.59  \\

$\mathcal{P} + \mathcal{M} + \mathcal{T}  \rightarrow   \mathcal{P} + \mathcal{T}$ & 67.88 & 60.58 & 69.04 & 76.34  \\

$\mathcal{P} +  \mathcal{T}  \rightarrow   \mathcal{P} + \mathcal{M} + \mathcal{T}$ & 68.53 & 60.81 & 70.14 & 78.37  \\

$\mathcal{M} + \mathcal{T} + \mathcal{R}  \rightarrow   \mathcal{T} + \mathcal{R}$ & 69.85 & 60.98 & 73.62 & 81.62  \\

$\mathcal{T} +  \mathcal{R}  \rightarrow   \mathcal{M} + \mathcal{T} + \mathcal{R}$ & 70.68 & 61.73 & 73.91 & 81.91  \\

$\mathcal{P} + \mathcal{M} + \mathcal{T} + \mathcal{R} \rightarrow  \mathcal{P} + \mathcal{T} + \mathcal{R}$ & 70.76 & 62.14 & 73.79 & 82.37  \\

$\mathcal{P} + \mathcal{M} + \mathcal{S} + \mathcal{T} + \mathcal{R} \rightarrow  \mathcal{P} + \mathcal{S}+ \mathcal{T} + \mathcal{R}$ & 70.76 & 62.14 & 74.08 & 82.26  \\

$\mathcal{P} + \mathcal{S} + \mathcal{T} + \mathcal{R} \rightarrow \mathcal{P} + \mathcal{M} + \mathcal{S}+ \mathcal{T} + \mathcal{R}$ & 72.84 & 64.11  & 76.75  & 84.58  \\

$\mathcal{P} + \mathcal{T} + \mathcal{R} \rightarrow \mathcal{P} + \mathcal{M} + \mathcal{T} + \mathcal{R}$ & \textbf{73.29} & \textbf{64.63}  & \textbf{77.10}  & \textbf{84.92}  \\

\arrayrulecolor{black!10}\midrule\arrayrulecolor{black}
\multicolumn{5}{l}{\textcolor{black!60}{\textit{3RScan predicted data}}} \\[-0.2ex] %
$\mathcal{P} + \mathcal{S} +  \mathcal{T} + \mathcal{R} \rightarrow \mathcal{P} + \mathcal{T} + \mathcal{R}$ & 39.54 & 20.97  &  48.47  & 63.00  \\

$\mathcal{P} + \mathcal{S} + \mathcal{T} +  \mathcal{R}  \rightarrow   \mathcal{T} + \mathcal{R}$ & 44.01 & 25.51 & 52.65 & 67.39  \\

$\mathcal{P} + \mathcal{T} + \mathcal{R} \rightarrow \mathcal{T} + \mathcal{R}$ &50.50 & 32.10 & 61.16  & 73.91  \\
$\mathcal{P} + \mathcal{S} + \mathcal{T} + \mathcal{R} \rightarrow \mathcal{S} + \mathcal{T} + \mathcal{R}$ & \textbf{52.51} & \textbf{34.94} & \textbf{69.31}  & \textbf{75.47}  \\

\bottomrule
\end{tabular}
}
\end{table}

\section{Conclusion}
We presented SGAligner++, a robust and lightweight framework for aligning 3D scene graphs across modalities using open-vocabulary language cues and learned joint embeddings. Unlike prior approaches, SGAligner++ integrates geometry, semantics, and spatial relationships from diverse input modalities, enabling accurate and efficient alignment even under challenging conditions such as partial overlap, noise, and missing data. Through extensive evaluation on both annotated and predicted datasets, we showed that SGAligner++ consistently outperforms existing baselines in terms of accuracy, generalization, and scalability. Its efficient design and open-vocabulary grounding make it well-suited for real-world robotics applications requiring scene understanding and cross-modal alignment. Future work may explore further extensions to outdoor scenes, video streams, and interactive environments.

\bibliographystyle{./bib/IEEEtran} 
\bibliography{./bib/IEEEabrv,./bib/IEEEexample}

\begin{thebibliography}{10}
\providecommand{\url}[1]{#1}
\csname url@rmstyle\endcsname
\providecommand{\newblock}{\relax}
\providecommand{\bibinfo}[2]{#2}
\providecommand\BIBentrySTDinterwordspacing{\spaceskip=0pt\relax}
\providecommand\BIBentryALTinterwordstretchfactor{4}
\providecommand\BIBentryALTinterwordspacing{\spaceskip=\fontdimen2\font plus
\BIBentryALTinterwordstretchfactor\fontdimen3\font minus \fontdimen4\font\relax}
\providecommand\BIBforeignlanguage[2]{{%
\expandafter\ifx\csname l@#1\endcsname\relax
\typeout{** WARNING: IEEEtran.bst: No hyphenation pattern has been}%
\typeout{** loaded for the language `#1'. Using the pattern for}%
\typeout{** the default language instead.}%
\else
\language=\csname l@#1\endcsname
\fi
#2}}

\bibitem{Koch_2024_WACV}
S.~Koch, P.~Hermosilla, N.~Vaskevicius, M.~Colosi, and T.~Ropinski, ``Sgrec3d: Self-supervised 3d scene graph learning via object-level scene reconstruction,'' in \emph{IEEE/CVF Winter Conference on Applications of Computer Vision (WACV)}, January 2024.

\bibitem{hughes2022hydra}
N.~Hughes, Y.~Chang, and L.~Carlone, ``{Hydra: A real-time spatial perception system for 3D scene graph construction and optimization},'' \emph{arXiv preprint arXiv:2201.13360}, 2022.

\bibitem{sarkar2025crossover}
S.~D. Sarkar, O.~Miksik, M.~Pollefeys, D.~Barath, and I.~Armeni, ``Crossover: 3d scene cross-modal alignment,'' in \emph{The IEEE Conference on Computer Vision and Pattern Recognition (CVPR)}, 2025.

\bibitem{miao2024scenegraphloc}
Y.~Miao, F.~Engelmann, O.~Vysotska, F.~Tombari, M.~Pollefeys, and D.~B. Bar{\'a}th, ``{SceneGraphLoc: Cross-Modal Coarse Visual Localization on 3D Scene Graphs},'' in \emph{European Conference on Computer Vision (ECCV)}, 2024.

\bibitem{Rosinol21}
A.~Rosinol, A.~Violette, M.~Abate, N.~Hughes, Y.~Chang, J.~Shi, A.~Gupta, and L.~Carlone, ``{Kimera: From SLAM to spatial perception with 3D dynamic scene graphs},'' \emph{IJRR}, 2021.

\bibitem{agia2022taskography}
C.~Agia, K.~Jatavallabhula, M.~Khodeir, O.~Miksik, V.~Vineet, M.~Mukadam, L.~Paull, and F.~Shkurti, ``Taskography: Evaluating robot task planning over large 3d scene graphs,'' in \emph{CoRL}, 2022.

\bibitem{armeni20193d}
I.~Armeni, Z.-Y. He, J.~Gwak, A.~R. Zamir, M.~Fischer, J.~Malik, and S.~Savarese, ``3d scene graph: A structure for unified semantics, 3d space, and camera,'' in \emph{Proceedings of the IEEE/CVF international conference on computer vision}, 2019.

\bibitem{learn3dssg}
J.~Wald, H.~Dhamo, N.~Navab, and F.~Tombari, ``Learning 3d semantic scene graphs from 3d indoor reconstructions,'' \emph{CoRR}, 2020.

\bibitem{3dsg}
A.~Rosinol, A.~Gupta, M.~Abate, J.~Shi, and L.~Carlone, ``3d dynamic scene graphs: Actionable spatial perception with places, objects, and humans,'' \emph{CoRR}, 2020.

\bibitem{sarkar2023sgaligner}
S.~D. Sarkar, O.~Miksik, M.~Pollefeys, D.~Barath, and I.~Armeni, ``Sgaligner: 3d scene alignment with scene graphs,'' in \emph{Proceedings of the IEEE/CVF International Conference on Computer Vision (ICCV)}, October 2023.

\bibitem{sgpgm}
Y.~Xie, A.~Pagani, and D.~Stricker, ``Sg-pgm: Partial graph matching network with semantic geometric fusion for 3d scene graph alignment and its downstream tasks,'' in \emph{Proceedings of the IEEE/CVF Conference on Computer Vision and Pattern Recognition (CVPR)}, June 2024.

\bibitem{Liu2025SGRegGA}
C.~Liu, Z.~Qiao, J.~Shi, K.~Wang, P.~Liu, and S.~Shen, ``Sg-reg: Generalizable and efficient scene graph registration,'' \emph{ArXiv}, 2025.

\bibitem{MultiKE}
Q.~Zhang, Z.~Sun, W.~Hu, M.~Chen, L.~Guo, and Y.~Qu, ``Multi-view knowledge graph embedding for entity alignment,'' in \emph{IJCAI}, 2019, pp. 5429--5435.

\bibitem{eva}
F.~Liu, M.~Chen, D.~Roth, and N.~Collier, ``Visual pivoting for (unsupervised) entity alignment,'' in \emph{Proceedings of the AAAI Conference on Artificial Intelligence}, vol.~35, no.~5, 2021.

\bibitem{lin2022multi}
Z.~Lin, Z.~Zhang, M.~Wang, Y.~Shi, X.~Wu, and Y.~Zheng, ``Multi-modal contrastive representation learning for entity alignment,'' in \emph{Proceedings of the 29th International Conference on Computational Linguistics}, 2022.

\bibitem{sarlin20superglue}
P.-E. Sarlin, D.~DeTone, T.~Malisiewicz, and A.~Rabinovich, ``{SuperGlue}: Learning feature matching with graph neural networks,'' in \emph{CVPR}, 2020.

\bibitem{whereami}
J.~Chen, D.~Barath, I.~Armeni, M.~Pollefeys, and H.~Blum, ````where am i?'' scene retrieval with language,'' in \emph{Computer Vision – ECCV 2024: 18th European Conference, Milan, Italy, September 29–October 4, 2024, Proceedings, Part XXXVII}.\hskip 1em plus 0.5em minus 0.4em\relax Berlin, Heidelberg: Springer-Verlag, 2024.

\bibitem{ssg3d}
J.~Wald, H.~Dhamo, N.~Navab, and F.~Tombari, ``{Learning 3D Semantic Scene Graphs from 3D Indoor Reconstructions},'' in \emph{Proceedings IEEE Conference on Computer Vision and Pattern Recognition (CVPR)}, 2020.

\bibitem{jia2024sceneversescaling3dvisionlanguage}
B.~Jia, Y.~Chen, H.~Yu, Y.~Wang, X.~Niu, T.~Liu, Q.~Li, and S.~Huang, ``Sceneverse: Scaling 3d vision-language learning for grounded scene understanding,'' in \emph{European Conference on Computer Vision (ECCV)}, 2024.

\bibitem{qi2016pointnet}
C.~R. Qi, H.~Su, K.~Mo, and L.~J. Guibas, ``Pointnet: Deep learning on point sets for 3d classification and segmentation,'' \emph{arXiv preprint arXiv:1612.00593}, 2016.

\bibitem{veličković2018graphattentionnetworks}
P.~Veli{\v{c}}kovi{\'{c}}, G.~Cucurull, A.~Casanova, A.~Romero, P.~Li{\`{o}}, and Y.~Bengio, ``{Graph Attention Networks},'' \emph{International Conference on Learning Representations}, 2018.

\bibitem{blip2}
J.~Li, D.~Li, S.~Savarese, and S.~Hoi, ``Blip-2: bootstrapping language-image pre-training with frozen image encoders and large language models,'' in \emph{Proceedings of the 40th International Conference on Machine Learning}, 2023.

\bibitem{homeomulti}
R.~Cipolla, Y.~Gal, and A.~Kendall, ``Multi-task learning using uncertainty to weigh losses for scene geometry and semantics,'' in \emph{2018 IEEE/CVF Conference on Computer Vision and Pattern Recognition}, 2018.

\bibitem{scannet}
A.~Dai, A.~X. Chang, M.~Savva, M.~Halber, T.~Funkhouser, and M.~Nie{\ss}ner, ``Scannet: Richly-annotated 3d reconstructions of indoor scenes,'' in \emph{Proc. Computer Vision and Pattern Recognition (CVPR), IEEE}, 2017.

\bibitem{3rscan}
J.~Wald, A.~Avetisyan, N.~Navab, F.~Tombari, and M.~Nie{\ss}ner, ``Rio: 3d object instance re-localization in changing indoor environments,'' in \emph{Proceedings of the IEEE/CVF International Conference on Computer Vision}, 2019.

\bibitem{scannotate}
S.~Ainetter, S.~Stekovic, F.~Fraundorfer, and V.~Lepetit, ``Automatically annotating indoor images with cad models via rgb-d scans,'' in \emph{Proceedings of the IEEE/CVF Winter Conference on Applications of Computer Vision}, 2023.

\bibitem{wu2021scenegraphfusionincremental3dscene}
S.-C. Wu, J.~Wald, K.~Tateno, N.~Navab, and F.~Tombari, ``{SceneGraphFusion: Incremental 3D Scene Graph Prediction from RGB-D Sequences},'' in \emph{Proceedings IEEE Conference on Computer Vision and Pattern Recognition (CVPR)}, 2021.

\bibitem{geotr}
Z.~Qin, H.~Yu, C.~Wang, Y.~Guo, Y.~Peng, and K.~Xu, ``Geometric transformer for fast and robust point cloud registration,'' in \emph{Proceedings of the IEEE/CVF Conference on Computer Vision and Pattern Recognition (CVPR)}, June 2022.

\bibitem{predator}
S.~Huang, Z.~Gojcic, M.~Usvyatsov, A.~Wieser, and K.~Schindler, ``Predator: Registration of 3d point clouds with low overlap,'' in \emph{Proceedings of the IEEE/CVF Conference on Computer Vision and Pattern Recognition (CVPR)}, June 2021.

\end{thebibliography}

\end{document}